\documentclass[10pt]{article}
\usepackage[preprint]{tmlr}
\usepackage[letterpaper,left=1in,right=1in,top=1.04in,bottom=1.1in,includefoot]{geometry}
\renewcommand{\headrulewidth}{0pt}

\usepackage{hyperref}
\hypersetup{hidelinks,
  pdftitle={Luck Is Not Skill: When Do Paired Rollouts Help Group-Relative RL of LLM Agents?},
  pdfauthor={Nazmus Sakib}}
\usepackage{url}
\usepackage{graphicx}
\usepackage{booktabs}
\usepackage{amsmath}
\usepackage{amssymb}
\usepackage{amsthm}
\usepackage{multirow}
\usepackage{array}
\usepackage{tabularx}
\usepackage{placeins}

\newtheorem{proposition}{Proposition}

\newcommand{\E}{\mathbb{E}}
\newcommand{\Var}{\mathrm{Var}}
\newcommand{\Cov}{\mathrm{Cov}}
\newcommand{\Tr}{\mathrm{Tr}}
\newcommand{\Rbar}{\bar{R}}
\newcommand{\Sbar}{\bar{S}}
\newcommand{\gc}{g_{\mathrm{c}}}
\newcommand{\gpair}{g_{\mathrm{paired}}}
\newcommand{\gind}{g_{\mathrm{indep}}}
\newcommand{\senv}{\sigma^2_{\mathrm{env}}}
\newcommand{\spol}{\sigma^2_{\mathrm{pol}}}

\newcommand{\cTwoAucMean}{+0.037}
\newcommand{\cTwoAucCI}{[-0.004, +0.078]}
\newcommand{\cTwoFinalMean}{+5.1}
\newcommand{\hThreeMean}{+5.4}
\newcommand{\hThreeCI}{[-1.0, +12.2]}
\newcommand{\hFiveMean}{+2.3}
\newcommand{\hFiveLower}{-0.4}
\newcommand{\hFiveUpper}{+5.2}

\newcommand{\cFourAucMean}{+0.003}
\newcommand{\cFourAucCI}{[-0.029, +0.033]}
\newcommand{\cFourFinalPaired}{89.4}
\newcommand{\cFourFinalIndep}{88.0}
\newcommand{\cZeroFinalClean}{89.9}

\newcommand{\probeOutcomeRedMin}{21}\newcommand{\probeOutcomeRedMax}{30}

\newcommand{\probeTransRedMin}{40}\newcommand{\probeTransRedMax}{63}
\newcommand{\probeLhsMin}{0.19}\newcommand{\probeLhsMax}{0.27}
\newcommand{\probeWithinOutMin}{29}\newcommand{\probeWithinOutMax}{37}
\newcommand{\probeWithinTransMin}{87}\newcommand{\probeWithinTransMax}{88}

\newcommand{\probeRhoZeroMin}{3.9}\newcommand{\probeRhoZeroMax}{7.2}
\newcommand{\probeRhoMeanMin}{0.33}\newcommand{\probeRhoMeanMax}{0.48}
\newcommand{\probeP}{33,638,400}

\title{Luck Is Not Skill: When Do Paired Rollouts Help\\Group-Relative RL of LLM Agents?}

\author{\name Nazmus Sakib \\
      \addr BRAC University, Dhaka, Bangladesh}

\begin{document}

\maketitle
\begingroup
\renewcommand{\thefootnote}{\fnsymbol{footnote}}
\footnotetext[1]{Language-model tools assisted with manuscript editing. The author verified every claim, number and reference and is responsible for the final content.}
\endgroup

\begin{abstract}
Group-relative reinforcement learning compares rollouts of the same prompt, but independent environment noise can obscure these comparisons. We study paired rollouts, which share an event-keyed noise schedule within each group while preserving each rollout's marginal distribution. Pairing removes the between-schedule component of reward-contrast variance, but need not reduce gradient variance. For one-sided grader noise, we derive an exact condition for reduction and give a counterexample in which reward contrasts improve while gradient variance increases. A controlled study trains a 2B tool-use agent under tool faults and grader flips, with three seeds per design. The protocol was registered with a disclosed, previously completed pilot. Under tool faults, pairing improves final noisy-test success by \cTwoFinalMean{} percentage points on average, with all three seed differences positive, but misses the registered learning-curve criterion. The criterion is also missed under grader flips: the validation-AUC difference is \cFourAucMean{} (95\% interval \cFourAucCI{}). A gradient probe on eight distinct checkpoints from two fault-trained trajectories finds lower mean-centered covariance traces under both noise types: \probeOutcomeRedMin{} to \probeOutcomeRedMax{}\% for grader flips and \probeTransRedMin{} to \probeTransRedMax{}\% for tool faults. These finite-sample measurements support the variance mechanism without establishing a general learning-speed benefit. The results distinguish improving reward comparisons, reducing estimator variance, and improving learning.

\end{abstract}

\section{Introduction}
\label{sec:intro}

Group-relative policy optimization (GRPO; \citealp{shao2024deepseekmath}) and related critic-free methods \citep{kool2019buy,ahmadian2024back,liu2025understanding,yu2025dapo} compare several rollouts of the same prompt. Each rollout receives an advantage based on its reward relative to the group, often normalized by the group's standard deviation. For tool-using agents, that comparison can reflect both policy behavior and environment randomness. Two otherwise similar trajectories may encounter different failures, rate limits, or grading errors.

We study \emph{paired rollouts}: all rollouts of a prompt share one noise schedule, while their policy samples remain independent. Different groups draw new schedules. This applies common random numbers, a classical simulation technique \citep{kahn1953methods,glasserman1992guidelines} also used in policy search \citep{ng2000pegasus,schulman2015trust}. Event keys align draws across trajectories that take different actions \citep{buffalo2026realizing}. The coupling preserves each rollout's marginal environment distribution and changes the dependence among siblings.

The effect on a group-relative gradient is not immediate. Under on-policy sampling, the leave-one-out estimator remains unbiased; the group-mean estimator retains its usual $(G-1)/G$ scaling. Pairing also removes the between-schedule component of reward-contrast variance. Neither fact guarantees lower gradient variance. We derive an exact condition under one-sided grader noise and construct a scalar bandit in which pairing reduces reward-contrast variance but increases gradient variance. The sign depends on the joint geometry of rewards and trajectory scores.

We test this distinction with a 2B tool-use agent in a back-office simulator. The study compares paired and independent schedules under tool faults and grader flips, with three seeds per design and matched training rollout budgets. Its registered protocol includes a disclosed pilot completed before the freeze. Both primary learning criteria are missed. Under tool faults, the average final noisy-test advantage is \cTwoFinalMean{} percentage points, positive on each seed; the registered recovery criterion also passes, although its bootstrap interval includes zero. Under grader flips, the learning-curve difference is small and uncertain.

A separate probe measures eight distinct checkpoints from two fault-trained trajectories. Mean-centered covariance traces are lower under pairing at every checkpoint for both noise types. Grader corruption is enumerated exactly conditional on sampled clean trajectories; tool faults are evaluated by regrouping a finite rollout pool. The probe contains no grader-trained checkpoints and does not establish that the measured variance reductions cause the learning differences. The contribution is a conditional estimator analysis with controlled evidence about both its benefits and its limits.

\section{Related work}
\label{sec:related}

\paragraph{Common random numbers.} Coupling the random draws of alternatives is a classical Monte Carlo technique \citep{kahn1953methods}; its benefits depend on the induced covariance \citep{glasserman1992guidelines}. PEGASUS fixes simulator randomness for policy search \citep{ng2000pegasus}, and TRPO's vine estimator uses common random numbers for action comparisons \citep{schulman2015trust}. Event-keyed hashing aligns draws when alternatives induce different event sequences \citep{buffalo2026realizing}. We analyze this coupling in a group-relative gradient, including a condition under which it helps and a counterexample to a general variance-reduction claim.

\paragraph{Group-relative estimators.} REINFORCE and policy-gradient theory provide the score-function framework \citep{williams1992simple,sutton2000policy}. Leave-one-out baselines reuse other samples of the same prompt \citep{kool2019buy,ahmadian2024back}. GRPO adds group standard-deviation normalization \citep{shao2024deepseekmath,guo2025deepseekr1}; Dr.~GRPO examines normalization biases \citep{liu2025understanding}; DAPO uses token-level loss aggregation \citep{yu2025dapo}. Our training uses group-standardized advantages with DAPO aggregation, without dynamic sampling. The theory concerns the mean-centered estimator; the probe also measures a standardized, importance-weighted estimator with a fixed token normalizer. Baseline analyses \citep{greensmith2004variance} motivate variance measurement, but changing a baseline and coupling environment draws are different interventions.

\paragraph{Noise and tool use.} Reward corruption has been studied through confusion-matrix models \citep{wang2020perturbed}, while spurious rewards can produce learning effects in some verifiable-reward settings \citep{shao2025spurious}. For tool-using agents, $\tau$-bench measures reliability across repeated trials \citep{yao2024taubench}, AgentNoiseBench evaluates controlled perturbations \citep{wang2026agentnoisebench}, and NoisyAgent combines noisy-environment training with a curriculum \citep{chen2026learning}. ToolRL studies rewards for tool learning \citep{qian2025toolrl}. Pairing instead changes the joint distribution of environment draws within a group while holding their marginal distribution fixed.

\paragraph{Implementation and registration.} Inference/training differences can make rollout sampling effectively off-policy; sequence-level correction and masking address aspects of this mismatch \citep{li2025trust}. We report the probe's weights and limit the on-policy theorem accordingly. Registration separates planned analyses from later interpretation \citep{nosek2018preregistration}. Our pilot reuse, amendment and deviations are disclosed in Section~\ref{sec:prereg} and Appendix~\ref{app:prereg}.

\section{Setting}
\label{sec:setting}

Fix a prompt $x$ and a policy $\pi_\theta$. An episode draws two independent sources of randomness: the environment's, $\omega$, which we call the \emph{schedule} (which tool calls fault, which reads go stale, whether the grader flips), and the policy's own sampling, $\xi$. A rollout is a trajectory $\tau = \tau(\omega, \xi)$ with reward $R = R(\omega, \xi)$ and score $S = \nabla_\theta \log \pi_\theta(\tau)$, the sum of the action log-probability gradients over the policy's own tokens; tokens returned by tools are not actions and do not enter $S$.

A group is $G$ rollouts of the same prompt with rewards $R_1, \dots, R_G$ and scores $S_1, \dots, S_G$. Group-relative methods score rollout $i$ by a contrast against its siblings,
\begin{equation}
A_i = R_i - \Rbar \quad \text{(mean-centered)}, \qquad A_i = \frac{R_i - \Rbar}{s + \epsilon} \quad \text{(standard-deviation normalized)},
\label{eq:advantages}
\end{equation}
with $\Rbar$ the group mean and $s$ the group standard deviation, and the group's gradient estimate is
\begin{equation}
g = \frac{1}{G} \sum_{i=1}^{G} A_i S_i .
\label{eq:g}
\end{equation}
Equation~\ref{eq:g} defines the estimator analyzed below. Training also uses advantage standardization, token-level loss aggregation and importance weights. The probe measures the mean-centered estimator and a fixed-normalizer approximation to the weighted loss-gradient contribution; neither includes the optimizer's subsequent transformation.

The two designs differ only in how the schedules are drawn. Under the \emph{independent} design, $\omega_1, \dots, \omega_G$ are independent draws: each rollout receives a separate environment draw. Under the \emph{paired} design, one $\omega$ is drawn per group and shared by its $G$ rollouts; different groups, and the same prompt at different steps, still draw different schedules. In both designs $\xi_1, \dots, \xi_G$ are independent, and each rollout's marginal law is the same.

\begin{figure}[t]
\centering\includegraphics[width=0.78\linewidth]{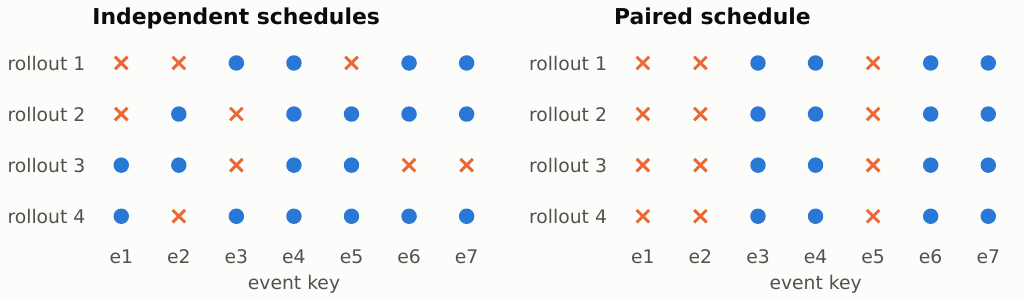}
\caption{Independent and paired schedules, illustrated for four rollouts and a common set of event keys. Marks represent draws attached to events, not a shared global call sequence. Under pairing, a given key receives the same draw; policy actions and history-dependent environment state can still change its consequences.}
\label{fig:design}
\end{figure}

\paragraph{Event-keyed schedules.} Figure~\ref{fig:design} illustrates the coupling. A schedule maps an event key to a pseudorandom draw. The key contains the tool, resource and repeat index of that tool and resource event; a separate episode key controls the grader. This follows the event-indexed construction of \citet{buffalo2026realizing}. Free-text arguments are excluded, so rewording does not select a new outage key. Equal keys and schedules give equal draws, but observations can differ because active rate limits, prior writes and persistent outages depend on each rollout's history. The coupling aligns exogenous randomness rather than forcing identical trajectories.

\section{Theory: from reward contrasts to gradient variance}
\label{sec:theory}

Proofs are in Appendix~\ref{app:proofs}. Assume $G>1$, independent policy samples conditional on the schedule, a schedule law and reward function with no direct dependence on $\theta$, finite second moments, and regularity for interchanging differentiation and expectation. The score identities additionally assume sampling from the policy whose log probabilities are differentiated. Throughout, $J(\theta) = \E_{\omega, \xi}[R]$ is the objective and $\gc$ denotes a group gradient computed from noise-free rewards.

\subsection{Unbiasedness}

\begin{proposition}[Unbiasedness under pairing]
\label{prop:unbiased}
For the leave-one-out mean-centered estimator $g_{\mathrm{LOO}} = \frac{1}{G}\sum_i \big(R_i - \frac{1}{G-1}\sum_{j \neq i} R_j\big) S_i$, $\E[g_{\mathrm{LOO}}] = \nabla_\theta J(\theta)$ under both designs. For the mean-including-self estimator of Equation~\ref{eq:g} the expectation is $\frac{G-1}{G}\nabla_\theta J(\theta)$ under both designs.
\end{proposition}

The reason is that $\omega$ enters the trajectory only through observations, while $S_i$ differentiates the policy's action probabilities alone: conditional on any fixed schedule and on the other rollouts, $\E[S_i \mid \omega, \xi_{-i}] = \sum_t \E[\nabla_\theta \log \pi_\theta(a_t \mid s_t) \mid s_t] = 0$ at every step, so a baseline built from the siblings' rewards has zero expected product with $S_i$ even though pairing correlates it with $R_i$. The statement does not extend to the standard-deviation normalized estimator, whose denominator depends on all $G$ rewards including $R_i$; pairing changes the joint law of the rewards and therefore of the denominator, and the two designs' expected updates can differ (in the bandit of Section~\ref{sec:condition} they differ in the third decimal). Pairing thus introduces no additional bias for these on-policy estimators; only the leave-one-out form is unbiased without a finite-group correction. This result does not directly apply to unweighted scores under a different sampling implementation.

\subsection{Reward contrasts and the luck share}
\label{sec:contrast}

For a fixed prompt, split the reward variance into the part explained by the schedule and the rest,
\begin{equation}
\senv = \Var_\omega\big(\E_\xi[R \mid \omega]\big), \qquad \spol = \E_\omega\big(\Var_\xi[R \mid \omega]\big), \qquad \Var(R) = \senv + \spol,
\end{equation}
and, when the total variance is positive, call $\lambda = \senv / (\senv + \spol) \in [0, 1]$ the \emph{luck share}, an intraclass correlation. $\spol$ is everything the schedule does not explain, which includes the interaction between policy and environment (the same fault can be survived or not depending on what the policy does), so it is not pure skill.

\begin{proposition}[Reward contrasts]
\label{prop:contrast}
For any two siblings $i \neq j$, $\Var(R_i - R_j) = 2(\senv + \spol)$ under the independent design and $2\spol$ under the paired design.
\end{proposition}

Paired siblings share $\omega$, so $\Cov(R_i, R_j) = \senv$ and the covariance term of $\Var(R_i - R_j)$ cancels exactly the between-schedule part. For averages of independent contrast replicates, the variance ratio is $1-\lambda$. When $\lambda<1$, matching that variance requires $1/(1-\lambda)$ times as many independent-design replicates. This is a statement about reward contrasts and nothing else: it is not an effective sample size for estimating the unconditional reward mean (there the exchangeable-correlation formula $G/(1 + (G-1)\lambda)$ applies, and pairing makes things worse), and it is not an effective sample size for the policy gradient, whose variance is the subject of the next section.

\subsection{The exact condition under one-sided outcome noise}
\label{sec:condition}

The variance of $g$ depends on the joint law of advantages and scores, not on the reward contrasts alone. By the law of total variance,
\begin{equation}
\Var(g) = \E_\omega[\Var(g \mid \omega)] + \Var_\omega(\E[g \mid \omega]),
\end{equation}
where $\omega$ is the group's single schedule under the paired design and the $G$-tuple under the independent design. The conditioning variable differs across designs, so this decomposition gives no general ordering of either term. A shared schedule can suppress an entire group's gradient. We measure empirical traces for transition noise; for one-sided outcome noise the difference has a closed form.

Let the environment be clean apart from the grader. A true success $r_i = 1$ is observed as $R_i = Z_i r_i$ with $Z_i \sim \mathrm{Bernoulli}(1-q)$; a true failure is always observed as $0$. Under the independent design the $Z_i$ are independent; under the paired design $Z_i = Z$ is shared by the group. Write $\gc = \frac{1}{G}\sum_i (r_i - \bar r) S_i = \frac{1}{G}\sum_i r_i (S_i - \Sbar)$ for the clean group gradient, with $\Sbar$ the mean score.

\begin{proposition}[Exact condition]
\label{prop:condition}
With mean-centered advantages, $\gpair = Z \gc$, both estimators have expectation $(1-q)\E[\gc]$, and
\begin{equation}
\Tr\Var(\gpair) - \Tr\Var(\gind) = q(1-q)\left[\E\|\gc\|^2 - \frac{1}{G^2}\,\E\sum_i r_i \|S_i - \Sbar\|^2\right].
\label{eq:condition}
\end{equation}
For $0<q<1$, pairing strictly lowers the trace of the gradient covariance if and only if $\E\|\gc\|^2 < \frac{1}{G^2}\E\sum_i r_i\|S_i - \Sbar\|^2$.
\end{proposition}

The left side is the size of the clean group gradient (its squared mean plus its variance); the right side is the score spread of the truly successful rollouts. Two regimes make the condition concrete.

\paragraph{Orthogonal scores of equal norm.} If the $S_i$ are mutually orthogonal with $\|S_i\|^2 = s^2$, an illustrative geometry rather than an assumption about language models, then for a group with $k$ true successes $\|\gc\|^2 = s^2 k(G-k)/G^3$ and $\frac{1}{G^2}\sum_i r_i \|S_i - \Sbar\|^2 = s^2 k(G-1)/G^3$, so the condition reads $k \geq 2$: pairing lowers the gradient variance for every group with at least two true successes and ties at $k \leq 1$. For an all-correct group ($k = G$) the left side is exactly zero whatever the score geometry, so pairing never loses on such a group and wins strictly whenever the successful rollouts' scores are not all identical. These are the spurious-variance groups of Section~\ref{sec:spurious}. At $q=0$ or $q=1$, both designs tie regardless of geometry.

\paragraph{Aligned scalar scores.} Consider $a\sim\mathrm{Bernoulli}(\sigma(\theta))$ at $\theta=0$, true reward $r=a$, score $S=a-\tfrac12$, $G=8$ and $q=0.1$. Exact enumeration gives reward-contrast variance 0.495 (independent) versus 0.450 (paired), but mean-centered gradient variance 0.002615 versus 0.005845. Pairing reduces the first by 9\% and increases the second 2.2-fold. Here $\E\|\gc\|^2=0.04956$ exceeds the score-spread term 0.01367. Standardization also changes the two expected updates. Appendix~\ref{app:bandit} gives the full enumeration.

Pairing therefore has no general gradient-variance guarantee. Equation~\ref{eq:condition} identifies the relevant score geometry; the probe measures both sides on sampled policy rollouts.

\subsection{Spurious-variance groups}
\label{sec:spurious}

Under independent grader flips, a group whose $G$ rollouts are all truly correct has non-zero observed variance with probability
\begin{equation}
P_{\mathrm{spurious}}(q, G) = 1 - (1-q)^G - q^G,
\end{equation}
which is $0.15$ at $q = 0.02$, $0.34$ at $q = 0.05$ and $0.57$ at $q = 0.10$ for $G = 8$. Under the paired design the probability is $0$: either nobody is flipped or everybody is, and both are zero-variance groups with zero advantages. Using the sample standard deviation and $\epsilon=10^{-4}$, one flipped rollout among eight correct ones receives an advantage of $-0.875/(\sqrt{0.125} + 10^{-4}) = -2.47$ and each sibling $+0.35$; the observed differences are entirely due to grading noise, yet the estimator penalizes a correct trajectory. For the unnormalized estimator these groups add variance of exactly the amount of Equation~\ref{eq:condition} in the all-correct case and do not bias the gradient, since the expected noisy reward is $(1-q)$ times the clean one; normalization can also change the expected update when averaged over general clean groups. The rate of spurious groups is a mechanism check, not a prediction of learning-curve magnitudes.

\subsection{What pairing does not do}

Pairing leaves the conditional policy-sampling law unchanged and cannot couple randomness outside the trainer's control. A batch of $B$ groups exposes pairing to $B$ schedules instead of $BG$, creating a possible diversity cost. We test a limited version of this concern on two held-out fault types. A lower covariance trace also does not establish a better optimization step: changes in mean gradients, normalization, importance weights and the optimizer can matter.

\section{Environment and pre-registered protocol}
\label{sec:protocol}

\subsection{A back-office tool environment}

The environment is a simulated back office over customers, orders, products and support tickets, exposed to the agent through thirteen tools: five reads (search customers, get customer, list orders, get order, check inventory), six writes (update shipping address, cancel order, issue refund, reserve stock, schedule shipment, create ticket), a \texttt{wait} tool for rate limits and \texttt{finish}. A task is a customer request in natural language composed of one to three sub-requests drawn from eight templates (address changes, cancellations with and without refund, partial refunds, reservations and shipments, tickets), each with a planned sequence of tool calls. The grader compares the final world state with the expected one field by field; this deterministic comparison defines the simulator's true-success metric, recorded separately from corrupted rewards. An episode may make at most $7 + 2n$ tool calls, $n$ being the number of planned writes, capped at 21; calls beyond the budget are refused and the episode ends. A scripted reference agent that retries and waits as needed solves the clean tasks in 6.6 calls on average and never exceeds the budget. Task pools are generated once from a fixed seed: 2{,}000 training tasks, a validation pool of 300 tasks (disjoint from training by customer pool and world seed), a separate test pool of 300 tasks used for final evaluation, and 16 diagnostic tasks with two to four writes for the luck-share design. Appendix~\ref{app:env} lists the tools, templates and fault types.

\subsection{Noise}

\emph{Transition noise} at rate $p$ per eligible tool call is a fixed mixture of five fault types: transient failure (an error; a retry draws a fresh fate), rate limit (an error with a retry-after; every call fails until the wait is over), outage (the same logical request fails for the rest of the episode), stale read (a read can return a snapshot from before the preceding write) and truncation (a list response is cut and must be paginated), with weights 0.45, 0.15, 0.10, 0.15 and 0.15, renormalized over the types applicable to each tool. The mixture was fixed at calibration, before the pre-registration was frozen, and never changed. Two further types, timeout after commit (a write is applied but reported as a timeout) and field dropout (a read loses one field), are never used in training and serve the held-out-type evaluation. At $p = 0.25$ about three quarters of the episodes of the current policies meet at least one fault; $p$ is a per-call rate, not an episode-failure rate. \emph{Outcome noise} at rate $q$ flips the reported grade of a correct end state to a failure; the true outcome is recorded alongside.

Schedules are the event-keyed streams of Section~\ref{sec:setting}. Evaluation schedules are seeded per task and per schedule index, identical across arms and checkpoints, so that exogenous draws are matched, although actions and resulting states may differ.

\paragraph{Luck-share diagnostic.} For a fixed policy and task, $K = 8$ schedules times $M = 8$ policy samples per schedule give an $8 \times 8$ table of rewards from which $\senv$ and $\spol$ are estimated by one-way random effects ($\senv = \max(0, (\mathrm{MS}_{\mathrm{between}} - \mathrm{MS}_{\mathrm{within}})/M)$, $\spol = \mathrm{MS}_{\mathrm{within}}$), and $\lambda$ is averaged over the 16 diagnostic tasks where it is defined, with a bootstrap interval over tasks. Tables are kept separate per checkpoint. Under the null the estimator has a clipping bias of about $0.03$ for $8 \times 8$ tables, which is reported rather than corrected.

\subsection{Training and conditions}

The policy is Qwen3.5-2B \citep{qwen2026qwen35}, a hybrid model whose linear-attention layers run on the kernels of \citet{yang2024fla}, adapted with LoRA \citep{hu2022lora} of rank 32 on all linear layers \citep{mangrulkar2022peft} and trained with the GRPO trainer of TRL \citep{vonwerra2020trl} with vLLM \citep{kwon2023efficient} generation in the same process. Every run uses 24 prompts $\times$ 8 rollouts per step for 100 steps (19{,}200 training episodes), one optimizer step per generation batch, standard-deviation normalized advantages, DAPO's token-level loss aggregation (the trainer default), learning rate $10^{-5}$, sampling temperature 1.0, no KL term, thinking disabled, and the tool-result tokens masked from the loss. A run is specified by its condition, arm and seed; seed $s$ of the paired arm and seed $s$ of the independent arm see the same training tasks in the same order.

\begin{table}[t]
\caption{The pre-registered tier-1 run matrix. Every paired-versus-independent comparison is a seed pair trained on one hardware and software stack.}
\label{tab:matrix}
\begin{center}
\begin{tabular}{l l l l l}
\toprule
condition & noise & arms & seeds & hypotheses \\
\midrule
C0 & none & clean & 0, 1 & reference curves \\
C2 & tool faults, $p = 0.25$, default mixture & paired, independent & 0, 1, 2 & H1a, H2a, H3, H5 \\
C4 & grader flips, $q = 0.10$ & paired, independent & 0, 1, 2 & H1a, H2b \\
\bottomrule
\end{tabular}
\end{center}
\end{table}

Table~\ref{tab:matrix} gives the run matrix: fourteen runs, each about 27{,}000 episodes including evaluations. The C2 paired seed-0 run is the gate-1 run that preceded the freeze; it has exactly the specification of C2 paired seed 0 (environment, data, schedules, trainer settings, evaluation protocol), and its inclusion was disclosed in the frozen document together with the fact that its validation curve was known when the threshold of the secondary endpoint was set. The decision rules were fixed before it finished and are applied to all seeds without change.

\subsection{Evaluation}

During training, every 20 steps, the policy is evaluated on 64 validation tasks under three regimes: clean, the at-training-noise regime (the C2 mixture at $p = 0.25$ for C2 and for the clean arm, grader flips at $q = 0.10$ for C4, whose learning metric is the true success on the clean set), and a \emph{matched challenge set} in which the first attempt of every write request fails transiently and nothing else does, identical for every policy and recoverable within budget. At the end, every policy is evaluated on the test pool (200 tasks $\times$ 4 schedules) under five regimes: clean, faults at $p = 0.10$ and $p = 0.25$, the held-out fault types at $p = 0.25$, and the matched challenge set, plus the at-training-noise regime when it is not one of those. The challenge uses one schedule per task, hence 200 episodes rather than 800. The learning metric is always the true, noise-free success. The primary endpoint of the learning hypotheses is the area under the validation curve (AUC), the trapezoidal mean of true success over the six evaluations, a number in $[0,1]$ with weights $(0.1,0.2,0.2,0.2,0.2,0.1)$ at steps $(0,20,40,60,80,100)$; a steps-to-threshold endpoint is secondary and censored rather than imputed. Intervals are percentile bootstrap intervals over 10{,}000 resamples that resample seeds as (paired, independent) pairs and tasks with all of a task's schedules together; episodes are never pooled as independent samples.

\subsection{Pre-registration}
\label{sec:prereg}

The protocol was frozen on 8 September 2026 and its hash timestamped by two RFC 3161 authorities \citep[for the rationale for registration, see][]{nosek2018preregistration}. Thirteen of fourteen included runs started after the freeze. The exception is the disclosed, completed gate-1 pilot. The final-test pool had also been inspected in calibration, gate 1 and a stack check; it is separate from training but not an untouched holdout. The hypotheses are as follows.

\begin{itemize}
\item \textbf{H1a (mechanism, counts).} Under grader flips, the independent arm's spurious-variance rate among all-correct training groups lies within $0.07$ of $0.57$ and the paired arm's is at most $0.02$, measured from the trainer's own groups. Under tool faults the zero-shot luck share is at least $0.15$ with a bootstrap lower bound above $0.05$, and, as a check of the implementation, every paired training group shares one resolved schedule seed while no independent group does.
\item \textbf{H1b (mechanism, gradients).} At the base model and at the checkpoints of two C2 paired trajectories, the gradient probe reports both sides of Equation~\ref{eq:condition} and the trace of the gradient covariance under both designs at matched rollout cost, for outcome noise at $q = 0.10$ (exact over masks conditional on the sampled clean groups) and for transition noise at $p = 0.25$ (resampled). H1b passes for a trajectory if $\Tr\Var(\gpair) < \Tr\Var(\gind)$ at every checkpoint for outcome noise and at a majority of checkpoints for transition noise. Whatever the outcome, both sides are reported.
\item \textbf{H2a, H2b (learning at matched rollout budget).} Seed $s$ of the paired arm against seed $s$ of the independent arm: the mean over seeds of the paired-minus-independent AUC difference is at least $0.03$ and every seed-level difference is positive; the final test success in the at-training-noise regime must not contradict the direction. H2a is the claim under tool faults (C2) and H2b under grader flips (C4, clean validation set); they are two scientific claims and neither substitutes for the other.
\item \textbf{H3 (recovery).} In C2, on the matched challenge set, the paired arm exceeds the independent arm by at least 5 points averaged over the step-40, 60 and 80 evaluations and over seeds, with every seed-level difference positive.
\item \textbf{H5 (held-out-type non-inferiority).} In C2, on the held-out fault types, the paired arm is non-inferior to the independent arm with a 3-point margin: the lower bound of the 90 percent bootstrap interval of the difference is above $-0.03$.
\end{itemize}

Registered predictions with confidences accompany the hypotheses (Table~\ref{tab:predictions} in Appendix~\ref{app:prereg}); the one about the probe, P13, predicted that the condition holds at the base model by a factor of at least two, that the outcome-noise ratio falls at later checkpoints, and that the transition-noise ratio is below one at a majority of checkpoints. Amendment A1 preceded the remaining training runs and probe measurements, but followed the known gate-1 results (Appendix~\ref{app:deviations}). The frozen provider assignment was not kept for every condition; each seed pair nevertheless trained on one accelerator type and software stack (Appendix~\ref{app:training}). Neither H2 criterion passes, so the paper adopts the estimator and boundary framing required by the registered gate-2 rule.

\section{Results}
\label{sec:results}

The tables and figures are generated from the run and probe summaries. Table~\ref{tab:outcomes} distinguishes the registered decision rules from evidence about effect size. In particular, a point-estimate rule can pass while its bootstrap interval includes zero.

\begin{table}[t]
\caption{Registered outcomes. Differences are paired minus independent. H1b uses the mean-centered estimator. Intervals are 95\% unless stated otherwise; pp denotes percentage points.}
\label{tab:outcomes}
\centering\small
\begin{tabularx}{\linewidth}{@{}l >{\raggedright\arraybackslash}X >{\raggedright\arraybackslash}X@{}}
\toprule
Hypothesis & Decision rule & Outcome \\
\midrule
H1a: mechanism & Initial reward luck share $\geq0.15$, lower bound $>0.05$; spurious rates near 0.57 and 0; schedule checks & Pass: luck share 0.33 to 0.38; spurious rates 0.558 and 0.000; schedule checks pass \\
H1b: probe & Lower trace at all outcome checkpoints and a majority of transition checkpoints, per trajectory & Pass: all eight distinct checkpoints under both noise types \\
H2a: C2 learning & Mean AUC difference $\geq0.03$, all seeds positive, consistent final direction & Miss: \cTwoAucMean{}; seed differences $+0.054,-0.002,+0.059$ \\
H2b: C4 learning & Same criterion, clean-validation AUC & Miss: \cFourAucMean{}; interval \cFourAucCI{} \\
H3: recovery & Challenge difference $\geq5$ pp at steps 40 to 80, all seeds positive & Pass by rule: \hThreeMean{} pp; interval \hThreeCI{} pp includes zero \\
H5: held-out types & 90\% lower bound above $-3$ pp & Pass: \hFiveMean{} pp; 90\% interval [\hFiveLower{},\hFiveUpper{}] pp \\
\bottomrule
\end{tabularx}
\end{table}

\subsection{Reward-level mechanism (H1a)}
\label{sec:mechanism}
At step 0, reward luck share under tool faults at $p=0.25$ is 0.33 to 0.38 across the six C2 runs; each task-bootstrap 95\% interval has a lower bound above 0.20. At step 100 it is 0.44 to 0.53. These estimates use 16 diagnostic tasks with $8\times8$ schedule by policy-sample tables and concern reward variance, not gradient variance. All 2{,}400 training groups per paired run share one resolved schedule seed; no independent group shares a single seed. Under grader flips, 2{,}080 of 3{,}726 all-correct independent groups have nonzero observed variance (55.8\%), versus 0 of 3{,}989 paired groups. The theoretical independent rate is 57.0\%. H1a passes.

Under tool faults, 52 to 54\% of paired groups have constant observed rewards, versus about 13\% of independent groups (Table~\ref{tab:groups}). A shared schedule can leave a whole group without a reward contrast. These counts do not determine whether the lost contrasts would have carried useful gradient information.

\subsection{Learning under tool faults (C2)}
\label{sec:c2}
\begin{figure}[t]
\centering\includegraphics[width=\linewidth]{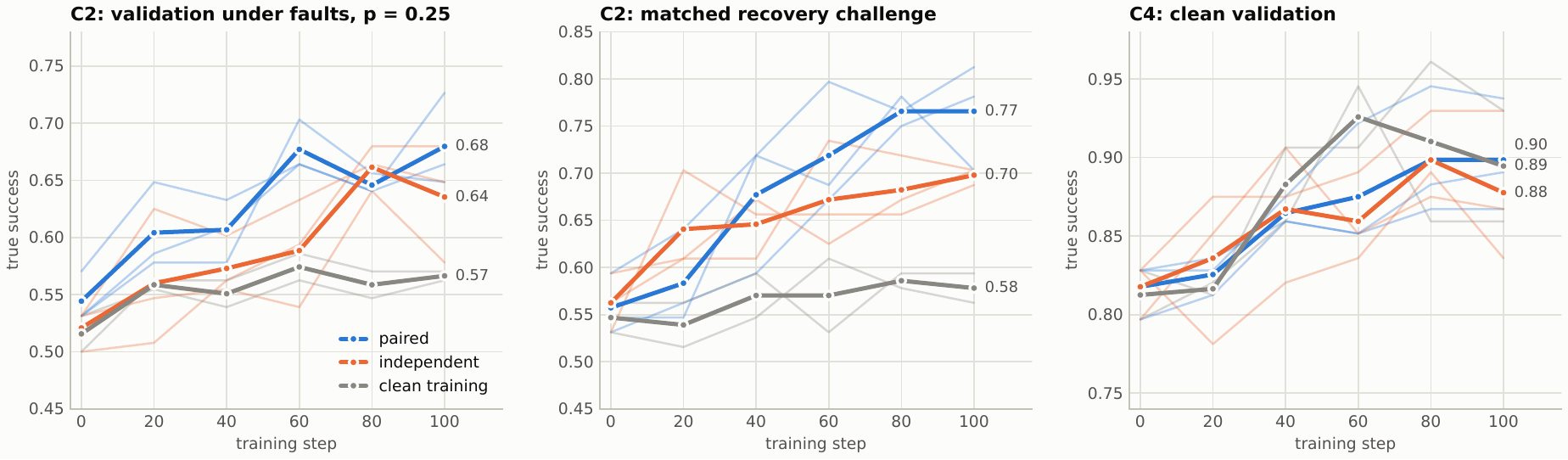}
\caption{Validation true success. C2 is evaluated under training-mixture faults and the matched challenge; C4 uses clean validation. Thin lines are seeds; bold lines are means. C0 is a two-seed clean-training reference. Each evaluation uses 64 tasks, with two schedules except for the challenge (one). Training rollout budgets are equal.}
\label{fig:curves}
\end{figure}

\begin{table}[t]
\caption{C2 primary endpoints by seed. AUC is the trapezoidal mean on noisy validation. Final success uses 200 test tasks with four schedules. AUC differences are raw units; final differences are percentage points. Secondary endpoints appear in Table~\ref{tab:c2_secondary}.}
\label{tab:c2}
\centering\small\begin{tabular}{@{}l rrr rrr@{}}
\toprule
 & \multicolumn{3}{c}{Validation AUC} & \multicolumn{3}{c}{Final success (\%)} \\
\cmidrule(lr){2-4}\cmidrule(lr){5-7}
Seed & paired & indep. & difference & paired & indep. & diff. (pp) \\
\midrule
0 & 0.641 & 0.587 & +0.054 & 65.4 & 61.4 & +4.0 \\
1 & 0.621 & 0.623 & -0.002 & 64.1 & 63.1 & +1.0 \\
2 & 0.626 & 0.567 & +0.059 & 71.0 & 60.8 & +10.2 \\
mean & 0.629 & 0.592 & +0.037 & 66.8 & 61.8 & +5.1 \\
\bottomrule
\end{tabular}

\end{table}

The validation-AUC difference is \cTwoAucMean{} (95\% interval \cTwoAucCI{}). Its mean exceeds 0.03, but seed 1 is negative, so H2a fails. The interval also includes zero. Seed-1 curves cross repeatedly (Table~\ref{tab:curves}). Final success under faults is 66.8\% versus 61.8\%, a mean difference of \cTwoFinalMean{} points, with individual differences $+4.0$, $+1.0$ and $+10.2$ points.

The challenge difference averaged over steps 40, 60 and 80 is \hThreeMean{} points, positive on each seed. H3 passes its registered point-estimate rule, but its 95\% interval, \hThreeCI{} points, includes zero. Final challenge success is 72.2\% versus 68.7\%; the final seed-1 challenge difference is negative. On the two held-out fault types, the mean difference is \hFiveMean{} points and the 90\% lower bound is \hFiveLower{}, above the $-3$-point margin. H5 passes for this evaluation; it does not rule out diversity costs under other fault distributions. Clean-trained C0 reaches 56.4\% on noisy test and 56.8\% on the challenge. These are descriptive references, not the matched H2a comparator.

\subsection{Learning under grader flips (C4)}
\label{sec:c4}
\begin{table}[t]
\caption{C4 primary endpoints. AUC uses clean validation and final success uses the clean test set. AUC differences are raw units; final differences are percentage points. C0 is a two-seed reference.}
\label{tab:c4}
\centering\small\begin{tabular}{@{}l rrr rrr@{}}
\toprule
 & \multicolumn{3}{c}{Validation AUC} & \multicolumn{3}{c}{Final success (\%)} \\
\cmidrule(lr){2-4}\cmidrule(lr){5-7}
Seed & paired & indep. & difference & paired & indep. & diff. (pp) \\
\midrule
0 & 0.891 & 0.890 & +0.001 & 90.5 & 90.2 & +0.3 \\
1 & 0.852 & 0.832 & +0.020 & 88.5 & 85.8 & +2.7 \\
2 & 0.850 & 0.863 & -0.013 & 89.1 & 87.9 & +1.2 \\
mean & 0.864 & 0.862 & +0.003 & 89.4 & 88.0 & +1.4 \\
\midrule
C0 reference & 0.878 & -- & -- & 89.9 & -- & -- \\
\bottomrule
\end{tabular}

\end{table}

The clean-validation AUC difference is \cFourAucMean{} (95\% interval \cFourAucCI{}). H2b fails the mean-margin and all-positive requirements. This is inconclusive about a learning benefit, not evidence of an exactly zero effect. Final clean-test success is \cFourFinalPaired{}\% for pairing, \cFourFinalIndep{}\% for independent schedules and \cZeroFinalClean{}\% for C0. These final percentages do not bound an AUC effect: the endpoint, seed count and training stacks differ. Grader-trained policies also remain less successful under tool faults (Table~\ref{tab:finals}), which are absent from their training distribution.

\subsection{Gradient probe (H1b)}
\label{sec:probe}
\paragraph{Design.} Each policy is frozen and both designs are compared on the same rollout pool. Trajectory B is C2 paired seed 1 at steps 0, 20, 40, 60, 80 and 100. Trajectory A is C2 paired seed 0 at steps 0, 80 and 100; earlier adapters were not retained. Step 0 is shared, giving eight distinct checkpoints, not eight independent training replicates. No C4-trained policies are probed.

The outcome experiment generates eight clean rollouts for each of 64 validation tasks. The transition experiment generates eight rollouts under each of eight schedules for each of 16 diagnostic tasks at $p=0.25$. Scores are differentiated over \probeP{} LoRA coordinates with the training tool mask. Inner products retain the full score vectors, with double-precision accumulation and no projection. This avoids sketching error, not finite-precision or sampling error. Generation and scoring use different implementations, so MC is a descriptive score-weighted estimator on these samples. The conditional grader identity still applies to its fixed scores; the on-policy unbiasedness proposition does not transfer automatically.

The primary estimator is mean-centered (MC), as in Equation~\ref{eq:g}. A secondary estimator uses standardized advantages and recorded importance weights, divided by a fixed reference token count per checkpoint (SW). It approximates a weighted loss-gradient contribution but omits variation from a random training-batch denominator, clipping and the optimizer (Appendix~\ref{app:probe}).

For outcome noise, corruption masks are enumerated exactly conditional on the sampled clean groups. Equation~\ref{eq:condition} has numerical residuals around $10^{-11}$ for traces of order $10^5$. For transitions, paired groups contain eight rollouts from one schedule. The independent comparator samples schedule labels with replacement and distinct rollouts within each chosen schedule, forming 64 groups per task. A stratified comparator uses one rollout per schedule. These overlapping groups reuse a finite pool and are not additional independent observations.

\begin{figure}[t]
\centering\includegraphics[width=\linewidth]{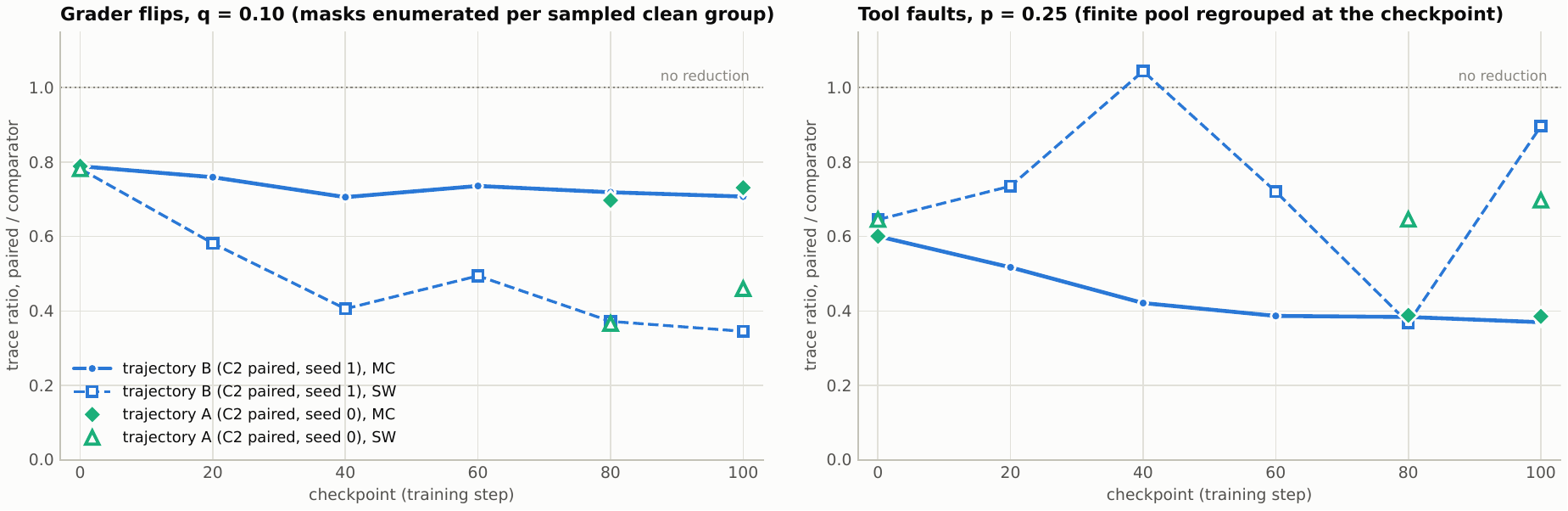}
\caption{Paired/independent covariance-trace ratios. MC is mean-centered; SW is standardized and importance-weighted with a fixed normalizer. Outcome expectations condition on sampled clean groups; transition estimates regroup the same finite pool. A and B share the base checkpoint. No uncertainty intervals are estimated for these ratios.}
\label{fig:probe}
\end{figure}
\begin{table}[t]
\caption{Probe ratios by checkpoint. $L/R$ compares the terms in Equation~\ref{eq:condition}, averaged over clean groups. Other entries are paired/comparator trace ratios. The primary transition comparator is independent resampling; ``strat.'' uses one rollout per schedule.}
\label{tab:probe}
\centering\small\begin{tabular}{@{}l r r rr rrr@{}}
\toprule
 & & & \multicolumn{2}{c}{Outcome} & \multicolumn{3}{c}{Transition} \\
\cmidrule(lr){4-5}\cmidrule(lr){6-8}
Traj. & step & $L/R$ & MC & SW & MC & SW & MC/strat. \\
\midrule
base & 0 & 0.275 & 0.789 & 0.782 & 0.601 & 0.645 & 0.552 \\
B & 20 & 0.243 & 0.759 & 0.582 & 0.517 & 0.735 & 0.464 \\
B & 40 & 0.196 & 0.706 & 0.406 & 0.421 & 1.044 & 0.389 \\
B & 60 & 0.220 & 0.736 & 0.494 & 0.387 & 0.721 & 0.343 \\
B & 80 & 0.206 & 0.719 & 0.372 & 0.384 & 0.368 & 0.340 \\
B & 100 & 0.197 & 0.707 & 0.345 & 0.370 & 0.897 & 0.332 \\
A & 80 & 0.189 & 0.697 & 0.366 & 0.388 & 0.646 & 0.347 \\
A & 100 & 0.216 & 0.731 & 0.460 & 0.385 & 0.697 & 0.354 \\
\bottomrule
\end{tabular}

\end{table}

\paragraph{Measured reduction.} The averaged condition holds at every checkpoint: $L/R$ ranges from \probeLhsMin{} to \probeLhsMax{}. MC traces fall by \probeOutcomeRedMin{} to \probeOutcomeRedMax{}\% under grader flips and \probeTransRedMin{} to \probeTransRedMax{}\% under tool faults. H1b passes on both trajectories. Final ratios are below the base ratios, although the outcome trend is not monotone. P13's factor-of-two base condition, endpoint direction and majority-of-transition-checkpoints predictions are met under this interpretation; a monotone improvement claim is not supported.

SW outcome ratios are below one at every checkpoint. Its transition ratio is below one at seven checkpoints, with B40 at 1.044. Thus 15 of 16 checkpoint and noise combinations favor pairing for SW. Both standardization and weighting differ from MC; without separate ablations, their contributions cannot be identified. The mean nonzero importance weight is \probeRhoMeanMin{} to \probeRhoMeanMax{}, with \probeRhoZeroMin{} to \probeRhoZeroMax{}\% masked to zero. These statistics describe the clean probe samples, not every training arm.

\subsection{Interpreting the decomposition}
\label{sec:reading}
Figure~\ref{fig:decomp} in Appendix~\ref{app:probe} shows the decomposition.

Under grader flips, conditional mask variance is \probeWithinOutMin{} to \probeWithinOutMax{}\% of the independent trace; pairing removes 73 to 81\% of it. The remainder is variability of conditional means across 64 sampled clean groups, one per task. It includes policy-sampling variability as well as task differences, rather than only differences among population task gradients. MC conditional means agree across grader designs, so this component is identical.

Under tool faults, the empirical within-task component is \probeWithinTransMin{} to \probeWithinTransMax{}\% of the independent trace. It mixes policy randomness, schedule randomness and their interaction; it is not an 87 to 88\% environmental contribution. Pooled trace reductions are also distinct from reductions in this within-task component.

The probe supports a local estimator effect, not a causal explanation of the learning difference between C2 and C4. It has no C4-trained checkpoints, uses different task sets for the two noise mechanisms, and changes the conditioning structure. Trajectory B has substantial probe reductions despite a negative H2a AUC difference. Whether these reductions improve optimization remains a question for further controlled tests.

\section{Discussion and limitations}
\label{sec:discussion}

\paragraph{Strength and scope.} The exact grader identity and counterexample support a conditional analysis of pairing. The probe finds lower mean-centered covariance traces at all sampled checkpoints. Learning evidence is weaker: both primary criteria fail, and the recovery difference's interval spans zero. The study covers one 2B model, one synthetic environment, 100 training steps and three seeds per comparison. The fault mixture was calibrated to produce appreciable reward luck share. Additional noise levels, outage-free training, blocking baselines, normalization ablations and an external benchmark were registered as later tiers but were not run. Three seeds give limited information about training-run variability, and pre-freeze test-pool exposure weakens the interpretation of final metrics as untouched confirmation.

\paragraph{Probe uncertainty.} Exact mask enumeration does not integrate over the population of trajectories. Transition estimates reuse finite rollout pools with overlapping groups; no fresh-rollout or task-bootstrap intervals accompany the trace ratios. Checkpoints share training trajectories and a base computation. Gradient norms depend on the LoRA parameterization and should not be transferred numerically to other adapters. Empirical mean-gradient differences and cosines also lack uncertainty estimates; we do not assume they are entirely sampling noise.

\paragraph{Importance weighting.} The trainer records a sequence ratio based on training/inference log-probability differences over policy tokens, masking ratios above 3. An exact, untruncated density ratio has expectation one under its sampling law, with suitable support. A nonpositive expected log-ratio does not imply a mean ratio below one. Masking and finite samples affect the observed mean. The reported clean-sample mean excludes zero weights; it neither gives the unconditional mean nor establishes a two- or three-fold change in gradient scale. Weights interact with advantages and scores, and can change the design comparison. The SW probe fixes the token normalizer and excludes optimizer transformations; a base-model smoke test is not validation of all checkpoints or full training batches.

\paragraph{Stack sensitivity.} Each seed pair trained on one accelerator type and software stack, but the C4 pairs did not all use the same one, departing from the frozen provider plan. An additional C4 paired seed-1 run on a second accelerator type obtained a clean-validation AUC of 0.902 against 0.852 for the run that entered the decision (Appendix~\ref{app:tables}). One cross-stack repetition does not isolate a hardware effect, but a difference of that size, larger than the mean C4 design difference, limits any comparison across stacks.

\paragraph{Practical implications.} Pairing is available where a simulator, replay layer or grader exposes controllable randomness. It preserves each rollout's marginal law and introduces no coupling bias to the on-policy leave-one-out estimator, but reduces schedule diversity and need not lower gradient variance. Reward luck share diagnoses reward contrasts; it is not a validated predictor of learning gains. At fixed policy, a trace ratio $r$ would correspond to a factor $1/r$ in the number of independent group replicates needed to match that trace, if the estimators have the same target mean. This variance calculation does not demonstrate a comparable improvement in training steps, tokens, wall time or final performance.

\section{Conclusion}
Sharing environment randomness improves within-group reward comparisons in a precise sense, but its effect on gradient variance depends on score geometry. We derive an exact condition for one-sided grader noise and exhibit a case where pairing increases gradient variance. On sampled tool-use checkpoints, pairing lowers mean-centered covariance traces under grader noise and tool faults. Learning results are more limited: final noisy-test differences favor pairing under tool faults, but neither registered primary learning criterion passes. Establishing when variance reduction improves optimization requires broader training comparisons and probes matched to those comparisons.

\subsubsection*{Code and Data Availability}
The code, the task pools, the frozen pre-registration with its timestamp tokens, and the run, decision and probe registers behind every table and figure are at \url{https://github.com/TheDeadcoder/paired-rollouts}; the numerical tables and the figure files are included with the arXiv source. Raw rollouts, trainer checkpoints and the probe's gradient evidence are not part of the repository.

\subsubsection*{Broader Impact Statement}
This study uses a synthetic environment without personal data or live service access. More reliable tool-use training could benefit downstream applications, but could also improve agents used for harmful purposes. The experiments do not establish deployment safety or a reduction in training compute. Applications involving consequential actions would still require authorization checks, monitoring, and evaluation under realistic failures.

\FloatBarrier

\bibliography{refs}
\bibliographystyle{tmlr}

\appendix
\FloatBarrier
\section{Proofs}
\label{app:proofs}

\paragraph{Proposition~\ref{prop:unbiased}.} Write $S_i = \sum_t \nabla_\theta \log \pi_\theta(a_t \mid s_t)$ over the policy's own actions in rollout $i$. The environment's randomness $\omega$ enters the trajectory only through the observations, so conditional on $\omega$ and on the other rollouts' sampling $\xi_{-i}$, the state $s_t$ at each step is a function of $\omega$ and of the actions before $t$, and $\E[\nabla_\theta \log \pi_\theta(a_t \mid s_t) \mid s_t] = \sum_a \pi_\theta(a \mid s_t)\nabla_\theta \log \pi_\theta(a \mid s_t) = \nabla_\theta \sum_a \pi_\theta(a \mid s_t) = 0$. Hence $\E[S_i \mid \omega, \xi_{-i}] = 0$ under both designs. The term $R_i S_i$ has expectation $\nabla_\theta \E[R]$ by the score-function identity applied to the trajectory distribution with $\omega$ integrated out (the transition kernel does not depend on $\theta$). For the leave-one-out baseline $b_i = \frac{1}{G-1}\sum_{j \neq i} R_j$, which is a function of $\omega$ and $\xi_{-i}$ only, $\E[b_i S_i] = \E[b_i \E[S_i \mid \omega, \xi_{-i}]] = 0$, so $\E[g_{\mathrm{LOO}}] = \frac{1}{G}\sum_i \nabla_\theta J = \nabla_\theta J$. For the mean-including-self estimator, $R_i - \Rbar = \frac{G-1}{G}(R_i - b_i)$, which gives the factor $\frac{G-1}{G}$. Sharing $\omega$ correlates $R_i$ with $b_i$ and changes nothing in this argument. \qed

\paragraph{Proposition~\ref{prop:contrast}.} $\Var(R_i - R_j) = \Var(R_i) + \Var(R_j) - 2\Cov(R_i, R_j)$ with $\Var(R_i) = \Var(R_j) = \senv + \spol$. Independent siblings are uncorrelated. Paired siblings share $\omega$ and are conditionally independent given it, so $\Cov(R_i, R_j) = \Cov(\E[R_i \mid \omega], \E[R_j \mid \omega]) = \Var_\omega(\E[R \mid \omega]) = \senv$. \qed

\paragraph{Proposition~\ref{prop:condition}.} With observed rewards $R_i = Z_i r_i$ and mean-centered advantages, $g = \frac{1}{G}\sum_i (R_i - \Rbar) S_i = \frac{1}{G}\sum_i R_i (S_i - \Sbar)$, since $\sum_i \Rbar S_i = \sum_i R_i \Sbar = G\Rbar\Sbar$. Under the paired design $Z_i = Z$, so $\gpair = Z \gc$ and, conditioning on the clean episode, $\E\|\gpair\|^2 = (1-q)\E\|\gc\|^2$. Under the independent design,
\[
\E\|\gind\|^2 = \frac{1}{G^2}\sum_{i,j} \E[Z_i Z_j]\,\E\big[r_i r_j \langle S_i - \Sbar, S_j - \Sbar\rangle\big]
= (1-q)^2\,\E\|\gc\|^2 + \big[(1-q) - (1-q)^2\big]\frac{1}{G^2}\E\sum_i r_i\|S_i - \Sbar\|^2 ,
\]
because the diagonal terms have $\E Z_i^2 = 1-q$ and the off-diagonal ones $(1-q)^2$, and $\|\gc\|^2 = \frac{1}{G^2}\sum_{i,j} r_i r_j \langle S_i - \Sbar, S_j - \Sbar\rangle$. Both designs have $\E g = (1-q)\E[\gc]$, so the squared means cancel in the difference of traces, which leaves $q(1-q)[\E\|\gc\|^2 - \frac{1}{G^2}\E\sum_i r_i\|S_i - \Sbar\|^2]$. \qed

\paragraph{Orthogonal scores.} With mutually orthogonal scores of squared norm $s^2$ and $k$ successes, $\|\gc\|^2 = \frac{s^2}{G^2}\sum_i (r_i - \bar r)^2 = \frac{s^2}{G^2}\Big[k\big(1 - \tfrac{k}{G}\big)^2 + (G-k)\big(\tfrac{k}{G}\big)^2\Big] = \frac{s^2 k (G-k)}{G^3}$, and $\|S_i - \Sbar\|^2 = s^2 - 2s^2/G + s^2/G = s^2 (G-1)/G$ for every $i$, so $\frac{1}{G^2}\sum_i r_i \|S_i - \Sbar\|^2 = \frac{s^2 k (G-1)}{G^3}$. The condition $k(G-k) < k(G-1)$ is $k \geq 2$, with equality at $k = 1$ and both sides zero at $k = 0$.

\subsection{The aligned-scalar bandit}
\label{app:bandit}

One-step bandit, $a \sim \mathrm{Bernoulli}(\sigma(\theta))$ at $\theta = 0$, true reward $r = a$, score $S = a - \tfrac12$, one group of $G = 8$, outcome noise $q = 0.1$. Enumerating the $2^8$ action tuples and, for each, the corruption masks exactly, gives: reward-contrast variance $0.4950$ (independent) and $0.4500$ (paired); mean-centered $\Tr\Var(g)$ $0.002615$ (independent) and $0.005845$ (paired) with expected update norm $0.1969$ under both; standard-deviation normalized (without inference importance weights) $\Tr\Var(g)$ $0.007053$ (independent) and $0.019724$ (paired) with expected update norms $0.3918$ and $0.3905$; $\E\|\gc\|^2 = 0.04956$ and $\frac{1}{G^2}\E\sum_i r_i\|S_i - \Sbar\|^2 = 0.01367$, whose difference times $q(1-q)$ is $0.00323$, the mean-centered gap. The same code path computes the probe's outcome-noise quantities, and reproducing these numbers is one of its unit tests.

\FloatBarrier
\section{Environment details}
\label{app:env}

\paragraph{World and tools.} The world holds customers (with search by name or email), orders with line items, shipping addresses and status (pending, paid, shipped, delivered, cancelled), products with inventory and reservations, shipments, and support tickets. Reads: \texttt{search\_customers}, \texttt{get\_customer}, \texttt{list\_orders}, \texttt{get\_order}, \texttt{check\_inventory}. Writes: \texttt{update\_shipping\_address}, \texttt{cancel\_order}, \texttt{issue\_refund}, \texttt{reserve\_stock}, \texttt{schedule\_shipment}, \texttt{create\_ticket}. Control: \texttt{wait} (advances a rate-limit clock) and \texttt{finish} (free; ends the episode with a summary). List responses are paginated with an offset.

\paragraph{Tasks.} A task is generated from a seed as one to three sub-requests (weights 20, 40 and 40 percent for one, two and three) drawn from eight templates with fixed weights: address change (3.0), cancel a pending order (2.0), cancel a paid order with refund (3.0), partial refund of a delivered order (3.0), reserve stock and ship (5.0), ship a reserved order (3.0), open a ticket about an order (1.2), open a ticket without an order (0.6). Each sub-request has a planned call sequence; the task's plan concatenates them and its expected end state is the world after the plan. The grader normalizes and compares the final snapshot with the expected one and reports success only when they agree field by field; it also reports whether the world changed at all. The call budget is $7 + 2n$, capped at 21, with $n$ the number of writes in the plan. Pools: training 2{,}000 tasks, validation 300, test 300 (generated after calibration with a disjoint seed range), diagnostic 16 (tasks with two to four writes drawn from the held-out generator distribution, disjoint from the exported validation set by world seed); the pools' SHA-256 hashes are recorded with every run and decision.

\paragraph{Fault types.} Transient failure: the call returns an error and the next attempt draws a fresh fate. Rate limit: the call returns an error with a retry-after; every call fails until the agent has waited that long. Outage: the same logical request (tool and resource) fails for the rest of the episode. Stale read: a read can return a snapshot from before the preceding write. Truncation: a list response is cut and requires pagination to complete. Held out for evaluation: timeout after commit (a write is applied but reported as a timeout) and field dropout (a read loses one field). Weights are renormalized over the types applicable to a tool (reads cannot time out after commit, writes cannot be stale).

\paragraph{Event keys and schedules.} The random draw for a call is a function of the schedule seed and the event key (tool name, resource identifier, repeat index of that event within the episode); free-text arguments such as reasons, summaries and search queries never enter the key. The grader's flip is one further draw per episode. Evaluation schedule $k$ of task $t$ uses the seed $\mathrm{eval\_schedule\_seed}(t, k)$, distinct across tasks and identical across arms and checkpoints; the outcome flip of an evaluation episode is a function of that seed, so at $q = 0.10$ about ten percent of evaluation episodes are flipped. In training, every row (a task at a step) carries a schedule seed derived from the run's seed and the row index; under the paired design all rollouts of the row share it, and under the independent design each rollout derives its own seed from the row's seed and its rollout slot.

\FloatBarrier
\section{Training and evaluation details}
\label{app:training}

\paragraph{Stack.} TRL 1.12.0 (\texttt{GRPOTrainer} with an environment factory for tool calling), vLLM 0.27.1 in colocated mode with 35 percent of the GPU memory, Transformers 5.16.1, PEFT 0.20.0, flash-linear-attention 0.5.2 for the Gated DeltaNet layers; PyTorch 2.13 with CUDA on the H100 accelerators and a ROCm build of PyTorch 2.11 on the MI300X accelerators. Generation is colocated with training, with weights synchronized for rollout generation.

\paragraph{Trainer settings.} Qwen3.5-2B in bfloat16, thinking disabled; LoRA rank 32, $\alpha = 64$, dropout 0, on all linear layers; 24 prompts per step, 8 rollouts per prompt, micro-batch 2 with 96 accumulation steps (one optimizer step per generation batch), 100 steps; learning rate $10^{-5}$; sampling temperature 1.0; maximum completion 6{,}144 tokens over at most 24 tool-calling turns, vLLM context 12{,}288; advantages standard-deviation normalized within the group; loss type DAPO (token-level aggregation over the generation batch); no KL term; tool-result tokens masked; vLLM importance-sampling correction at the trainer default (sequence-level mask, upper bound 3, no lower bound). Trainer state (adapter, optimizer, scheduler, RNG) is saved every 20 steps. Post-freeze C2 runs retain all five checkpoints; gate 1 retained only steps 80 and 100. A container that dies can be resumed from the last complete checkpoint with the previous attempt's evidence preserved, and every attempt is reported. Each of the fourteen selected runs has one recorded attempt; this does not count replaced infrastructure jobs as attempts of its replacement.

\paragraph{Evaluation.} Periodic (steps 0, 20, \dots, 100): 64 validation tasks $\times$ 2 schedules for the clean and the at-training-noise sets and 64 episodes of the matched challenge set. Final: 200 test tasks $\times$ 4 schedules per regime, 200 challenge episodes. Luck-share diagnostic: 16 tasks $\times$ 8 schedules $\times$ 8 samples at steps 0, 50 and 100 for seed 0 of every cell and at steps 0 and 100 for the other seeds. Training-group register: every training group's task, schedule seeds, true outcomes, observed rewards, advantages and zero-variance and spurious flags, taken from the trainer's own batches. AUC is the trapezoidal mean of true success over the six periodic evaluations. Intervals: percentile bootstrap, 10{,}000 resamples, seeds resampled as (paired, independent) pairs and tasks resampled with all their schedules together, 95 percent (90 percent for the non-inferiority bound).

\paragraph{Accelerators.} All C2 and C0 runs and the C4 seed-1 and seed-2 pairs trained on H100 accelerators with the CUDA stack; the C4 seed-0 pair trained on MI300X accelerators with the ROCm stack. Each seed pair trained on one accelerator type and one stack.

\FloatBarrier
\section{Pre-registration, amendment and deviations}
\label{app:prereg}

\paragraph{Freeze.} The pre-registration was frozen on 2026-09-08 (UTC) after gate 1 completed and before the remaining thirteen included runs; the SHA-256 of the frozen file was timestamped by two RFC 3161 authorities and the tokens are in the repository with the verification commands. The document fixes the thesis, definitions, the hypotheses and decision rules of Section~\ref{sec:prereg}, the numeric predictions of Table~\ref{tab:predictions}, the run matrix and provider assignment, the evaluation protocol, exclusion and relaunch rules, and the analysis plan. Its history (three drafts over the preceding week, including the withdrawal of an earlier general variance-reduction claim in favor of the exact condition) is part of the document.

\paragraph{Amendment A1 (before the remaining runs and probe).}
\label{app:deviations}
The frozen text placed the probe at the base model and at steps 20, 40, 60, 80 and 100 of the gate-1 run, but that run had kept only its checkpoints 80 and 100 (a checkpoint-retention limit at its training commit, known at the freeze and wrongly carried into the text). The amendment, written before the remaining training runs and probe outcomes, but after gate-1 results were known, defines trajectory A as gate 1 at the base model, 80 and 100 and trajectory B as C2 paired seed 1 at the base model and every 20 steps, applies the H1b rule to each trajectory over its available checkpoints, resamples the independent design as independent training would (schedule labels with replacement, then distinct rollouts within a schedule; the stratified design of the frozen text is reported alongside), adds a standardized, importance-weighted estimator and an exact conditional corruption average, and drops the random projection of the frozen text in favor of unprojected inner products over the selected LoRA coordinates. It affects H1b only.

\paragraph{Deviations and disclosures.} Before the freeze, three environment versions were built: the first calibration measured a luck share of $0.03$ (prediction P12 failed) and exposed defects (faults keyed by call index, a tool rejecting numeric postal codes, saturated diagnostic tasks); the second keyed faults by request and added the outage type and the call budget; the third, after an external review, keyed faults by tool and resource, gave every evaluation task its own schedules, added the test pool, the matched challenge set and the training-group register, and corrected the theory notes. All calibration registers are kept. The test pool's results were observed before the study in the zero-shot calibration, in the gate-1 run and in a stack check, so the frozen phrase ``never used for any decision'' is qualified to that exposure history. Four post-freeze C2 runs started two days after seed 0 because a planned second launch stage did not happen; nothing about the runs changed. The frozen plan assigned each condition to one provider; that assignment was not kept for C4 and C0: the C0 runs and two of the three C4 seed pairs trained on the H100 stack instead (Appendix~\ref{app:training}), each seed pair on one accelerator type and one stack. A completed C4 paired seed-1 run on the accelerator type that its independent counterpart did not use is reported separately as the cross-stack replication of Table~\ref{tab:crosshw} and is not counted as another seed. The probe's end-to-end smoke test failed once on a data-type bug in a checksum before the registered probes ran and was rerun under a new identifier after the fix; the registered probes ran once each. The deviation log does not contain contemporaneous entries for all of these later changes; we do not claim a complete prospective deviation record.

\paragraph{C4 final-direction endpoint.} The frozen rule for H2b substitutes the clean validation set for the at-training-noise set, and the decision script applies the same substitution to the final-direction check, which therefore uses final clean test success ($+1.4$ points on average; Table~\ref{tab:c4}). Read literally, the final-direction clause names the at-training-noise regime, which for C4 is the test set under grader flips at $q = 0.10$; the true-success differences there are $-0.1$, $+2.8$ and $+1.9$ percentage points (mean $+1.5$). The direction is the same under both readings, and H2b fails on the AUC and seed-sign requirements regardless. Both endpoints are reported and the original decision summary is retained in the repository's decision registers.

\begin{table}[htbp]
\caption{Registered numeric predictions and their outcomes. Predictions for unrun conditions (P3, P5, P7, P8, P10) are omitted; the frozen list has no P11. P12 originally failed before environment changes.}
\label{tab:predictions}
\begin{center}
\small
\begin{tabularx}{\linewidth}{@{}l X c X@{}}
\toprule
ID & Registered prediction & Conf. & Outcome \\
\midrule
P1 & independent spurious-variance rate $0.57 \pm 0.05$ at $q = 0.10$; paired below 0.02 & 0.90 & met: 0.558 and 0.000 \\
P2 & C4 AUC gap at least 0.03; paired final clean test success within 3 points of the clean run & 0.65 & missed on the gap (+0.003); met on the clean level (89.4 against 89.9) \\
P4 & C2 AUC gap between 0.02 and 0.06; final test gap at $p = 0.25$ between 2 and 8 points & 0.60 & met: +0.037 and +5.1 points \\
P6 & C2 challenge-set success at steps 40 to 80: paired ahead by at least 5 points & 0.55 & met: +5.4 points \\
P9 & held-out fault types: paired non-inferior with margin 3 points; point difference within 2 points & 0.80 & met on non-inferiority; the point difference is +2.3, outside 2 points \\
P12 & zero-shot luck share at $p = 0.25$ at least 0.15 & 0.60 & Original prediction failed in v1 ($0.030$). After environment changes, initial C2 values are $0.33$ to $0.38$. \\
P13 & at the base model lhs below rhs by a factor of at least 2; the outcome ratio falls with training; the transition ratio below 1 at a majority of checkpoints & 0.70 & Base $L/R=0.275$; final outcome ratios 0.707 (B), 0.731 (A), below base 0.789. All transition ratios below one. Outcome trend is not monotone. \\
\bottomrule
\end{tabularx}

\end{center}
\end{table}

\FloatBarrier
\section{Full result tables}
\label{app:tables}

\begin{table}[htbp]
\caption{C2 secondary endpoints. H3 is the paired-minus-independent challenge difference over steps 40, 60 and 80 in percentage points. Final success rates are percentages, with differences in percentage points.}
\label{tab:c2_secondary}
\centering\small\begin{tabular}{@{}l r rr rrr@{}}
\toprule
 & H3 & \multicolumn{2}{c}{Final challenge} & \multicolumn{3}{c}{Held-out types} \\
\cmidrule(lr){3-4}\cmidrule(lr){5-7}
Seed & diff. (pp) & paired & indep. & paired & indep. & diff. (pp) \\
\midrule
0 & +7.3 & 78.0 & 70.5 & 79.8 & 79.1 & +0.6 \\
1 & +1.6 & 62.5 & 72.0 & 77.5 & 76.5 & +1.0 \\
2 & +7.3 & 76.0 & 63.5 & 81.1 & 75.9 & +5.2 \\
mean & +5.4 & 72.2 & 68.7 & 79.5 & 77.2 & +2.3 \\
\bottomrule
\end{tabular}

\end{table}
\begin{figure}[htbp]
\centering\includegraphics[width=\linewidth]{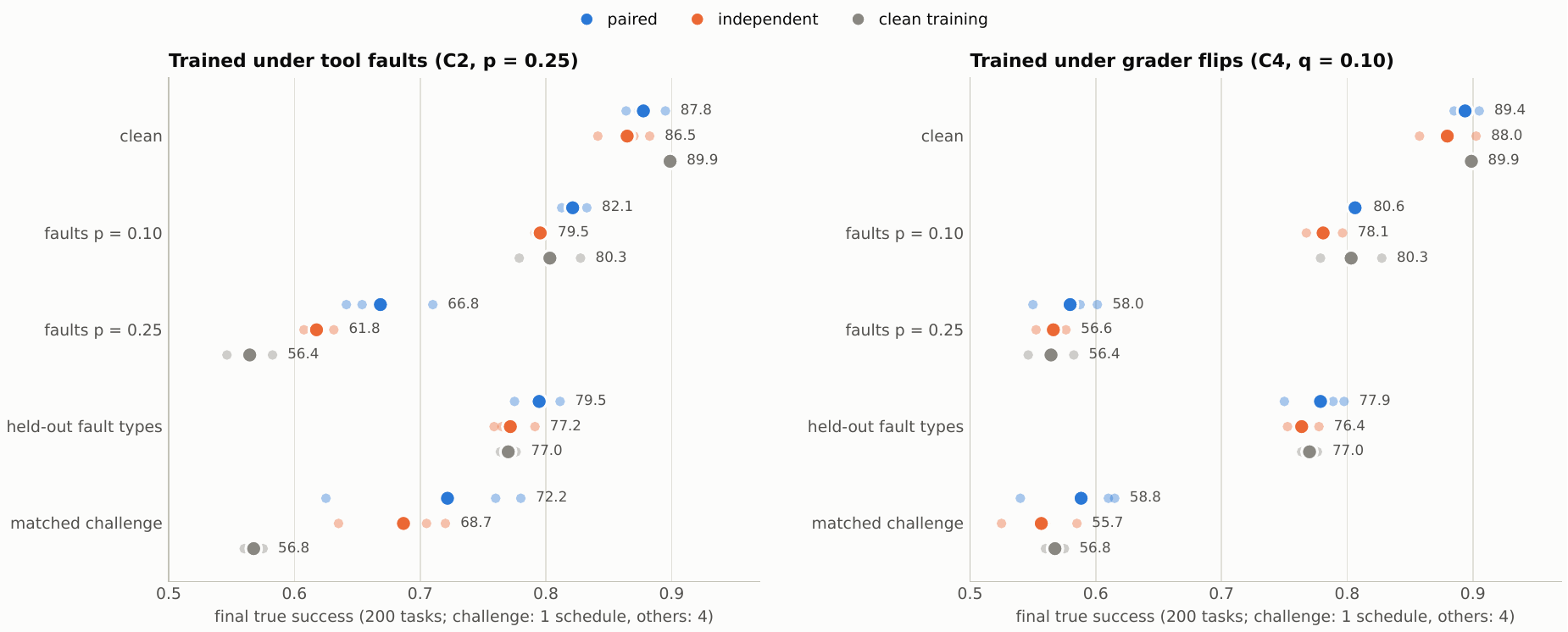}
\caption{Final true success. Small dots are seeds; large dots are means labelled in percent. Each regime uses 200 tasks with four schedules, except the challenge (one schedule). C0 has two seeds; other arms have three.}
\label{fig:final}
\end{figure}

\begin{table}[htbp]
\caption{Final test success (percent) per evaluation regime for every arm: individual runs and arm means. P: paired; I: independent. 200 tasks $\times$ 4 schedules per regime; 200 challenge episodes.}
\label{tab:finals}
\begin{center}
\small
\begin{tabular}{@{}l rrrrr@{}}
\toprule
Run & clean & $p=0.10$ & $p=0.25$ & held-out types & challenge \\
\midrule
C2 P 0 & 89.5 & 81.9 & 65.4 & 79.8 & 78.0 \\
C2 P 1 & 87.4 & 81.2 & 64.1 & 77.5 & 62.5 \\
C2 P 2 & 86.4 & 83.2 & 71.0 & 81.1 & 76.0 \\
C2 paired mean & 87.8 & 82.1 & 66.8 & 79.5 & 72.2 \\
\addlinespace
C2 I 0 & 87.0 & 79.8 & 61.4 & 79.1 & 70.5 \\
C2 I 1 & 88.2 & 79.8 & 63.1 & 76.5 & 72.0 \\
C2 I 2 & 84.1 & 79.1 & 60.8 & 75.9 & 63.5 \\
C2 independent mean & 86.5 & 79.5 & 61.8 & 77.2 & 68.7 \\
\addlinespace
C4 P 0 & 90.5 & 80.9 & 60.1 & 79.8 & 61.0 \\
C4 P 1 & 88.5 & 80.5 & 55.0 & 75.0 & 54.0 \\
C4 P 2 & 89.1 & 80.5 & 58.8 & 78.9 & 61.5 \\
C4 paired mean & 89.4 & 80.6 & 58.0 & 77.9 & 58.8 \\
\addlinespace
C4 I 0 & 90.2 & 79.6 & 57.0 & 75.2 & 52.5 \\
C4 I 1 & 85.8 & 76.8 & 55.2 & 76.1 & 58.5 \\
C4 I 2 & 87.9 & 77.9 & 57.6 & 77.8 & 56.0 \\
C4 independent mean & 88.0 & 78.1 & 56.6 & 76.4 & 55.7 \\
\addlinespace
C0 clean 0 & 90.0 & 77.9 & 58.2 & 76.4 & 56.0 \\
C0 clean 1 & 89.8 & 82.8 & 54.6 & 77.6 & 57.5 \\
C0 clean mean & 89.9 & 80.3 & 56.4 & 77.0 & 56.8 \\
\addlinespace
\bottomrule
\end{tabular}

\end{center}
\end{table}

\begin{table}[htbp]
\caption{Validation learning curves of every run (true success in percent at steps 0 to 100 on the at-training-noise set for C2 and on the clean set for C4 and C0) and their AUC.}
\label{tab:curves}
\begin{center}
\small
\begin{tabular}{@{}l l rrrrrrr@{}}
\toprule
Run & GPU & 0 & 20 & 40 & 60 & 80 & 100 & AUC \\
\midrule
C2 P 0 & H100 & 57.0 & 64.8 & 63.3 & 66.4 & 64.1 & 66.4 & 0.641 \\
C2 P 1 & H100 & 53.1 & 57.8 & 57.8 & 70.3 & 65.6 & 64.8 & 0.621 \\
C2 P 2 & H100 & 53.1 & 58.6 & 60.9 & 66.4 & 64.1 & 72.7 & 0.626 \\
C2 I 0 & H100 & 50.0 & 50.8 & 56.2 & 59.4 & 68.0 & 68.0 & 0.587 \\
C2 I 1 & H100 & 53.1 & 62.5 & 60.2 & 63.3 & 66.4 & 64.8 & 0.623 \\
C2 I 2 & H100 & 53.1 & 54.7 & 55.5 & 53.9 & 64.1 & 57.8 & 0.567 \\
C4 P 0 & MI300X & 82.8 & 82.8 & 87.5 & 92.2 & 94.5 & 93.8 & 0.891 \\
C4 P 1 & H100 & 82.8 & 83.6 & 85.9 & 85.2 & 86.7 & 86.7 & 0.852 \\
C4 P 2 & H100 & 79.7 & 81.2 & 85.9 & 85.2 & 88.3 & 89.1 & 0.850 \\
C4 I 0 & MI300X & 82.8 & 87.5 & 87.5 & 89.1 & 93.0 & 93.0 & 0.890 \\
C4 I 1 & H100 & 82.8 & 78.1 & 82.0 & 83.6 & 89.1 & 83.6 & 0.832 \\
C4 I 2 & H100 & 79.7 & 85.2 & 90.6 & 85.2 & 87.5 & 86.7 & 0.863 \\
C0 clean 0 & H100 & 79.7 & 82.0 & 85.9 & 94.5 & 85.9 & 85.9 & 0.863 \\
C0 clean 1 & H100 & 82.8 & 81.2 & 90.6 & 90.6 & 96.1 & 93.0 & 0.893 \\
\bottomrule
\end{tabular}

\end{center}
\end{table}

\begin{table}[htbp]
\caption{Training-group statistics from the trainer's own batches (2{,}400 groups per run) and the luck share of the reward from the diagnostic tables. The spurious rate is defined for C4 only (the fraction of all-correct groups with non-zero observed variance); the zero-variance fraction counts groups whose observed rewards are constant, which produce no gradient; $\max_i |A_i|$ is the group's largest normalized advantage, averaged over groups. Initial and final luck shares are shown; intermediate diagnostics remain in the run registers of the repository. These diagnostics use the tool-fault mixture for every training arm.}
\label{tab:groups}
\begin{center}
\small
\begin{tabular}{@{}l rrrrrr@{}}
\toprule
Run & all correct & spurious & constant $R$ & mean $\max|A|$ & $\lambda_0$ & $\lambda_{100}$ \\
\midrule
C2 P 0 & 726 & -- & 0.520 & 0.86 & 0.35 & 0.46 \\
C2 P 1 & 705 & -- & 0.517 & 0.86 & 0.38 & 0.52 \\
C2 P 2 & 791 & -- & 0.538 & 0.84 & 0.33 & 0.53 \\
C2 I 0 & 206 & -- & 0.127 & 1.40 & 0.35 & 0.44 \\
C2 I 1 & 231 & -- & 0.134 & 1.40 & 0.38 & 0.52 \\
C2 I 2 & 201 & -- & 0.135 & 1.37 & 0.33 & 0.51 \\
C4 P 0 & 1387 & 0.000 & 0.623 & 0.72 & 0.34 & 0.56 \\
C4 P 1 & 1245 & 0.000 & 0.570 & 0.81 & 0.41 & 0.61 \\
C4 P 2 & 1357 & 0.000 & 0.613 & 0.71 & 0.37 & 0.55 \\
C4 I 0 & 1355 & 0.574 & 0.243 & 1.39 & 0.34 & 0.54 \\
C4 I 1 & 1158 & 0.573 & 0.217 & 1.40 & 0.41 & 0.53 \\
C4 I 2 & 1213 & 0.526 & 0.245 & 1.35 & 0.37 & 0.45 \\
C0 clean 0 & 1323 & -- & 0.556 & 0.82 & 0.35 & 0.54 \\
C0 clean 1 & 1410 & -- & 0.590 & 0.76 & 0.38 & 0.63 \\
\bottomrule
\end{tabular}

\end{center}
\end{table}

\begin{table}[htbp]
\caption{Cross-stack replication of C4 paired seed 1: the same specification, seed, data and schedules on two accelerator types. The H100 run is the one that entered the decision, matching the accelerator type of its independent counterpart; the MI300X run is reported for comparison only.}
\label{tab:crosshw}
\begin{center}
\small
\begin{tabular}{l cccccc c c}
\toprule
GPU & 0 & 20 & 40 & 60 & 80 & 100 & AUC & final clean \\
\midrule
MI300X & 80.5 & 87.5 & 89.1 & 93.8 & 93.0 & 94.5 & 0.902 & 91.4 \\
H100 & 82.8 & 83.6 & 85.9 & 85.2 & 86.7 & 86.7 & 0.852 & 88.5 \\
\bottomrule
\end{tabular}
\end{center}
\end{table}

\FloatBarrier
\section{Gradient probe details}
\label{app:probe}

\begin{figure}[htbp]
\centering\includegraphics[width=\linewidth]{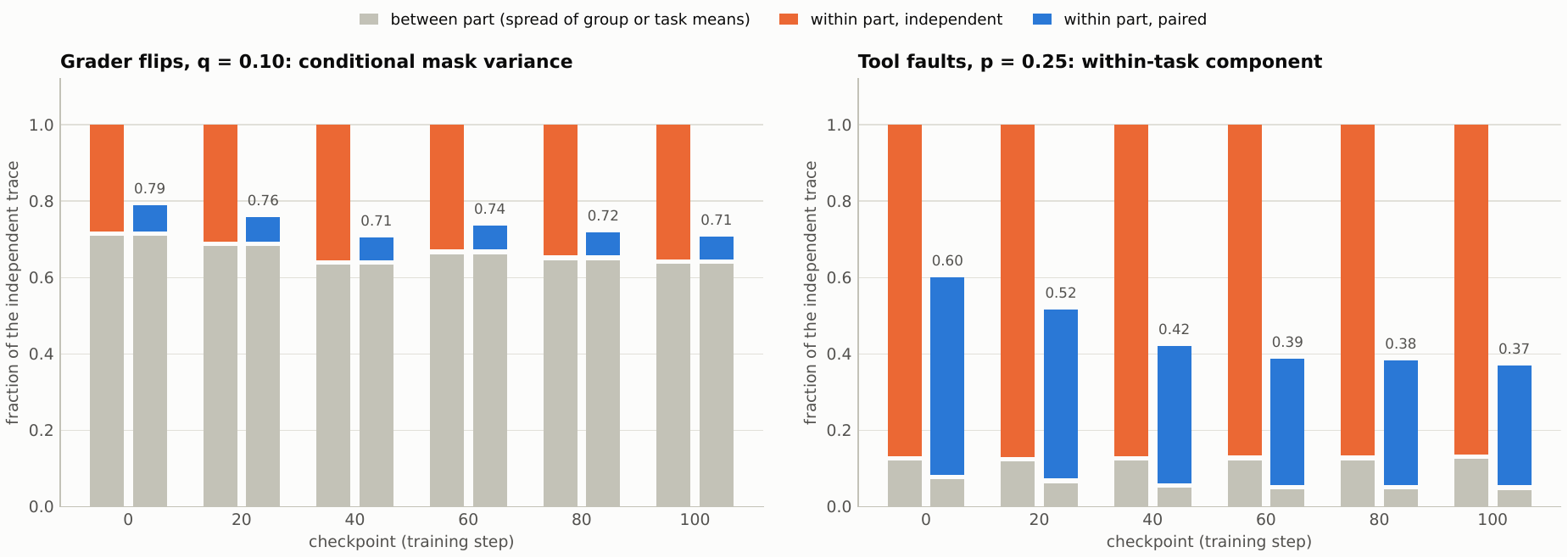}
\caption{MC trace decomposition for trajectory B, normalized by the independent total. Left: conditional grader-mask variance and the spread of conditional means across sampled clean groups. Right: empirical within-task and between-task transition components. Conditioning differs across panels; the colored components cannot both be interpreted as environment-only variance.}
\label{fig:decomp}
\end{figure}

\paragraph{Scores.} A saved adapter is loaded into the training model; the base computation uses a fresh seeded LoRA adapter. One backward pass identifies LoRA tensors whose gradient is present (\texttt{grad is not None}), including tensors whose gradient values are zero. The selected \probeP{} coordinates exclude vision adapters disconnected from text-only rollouts. Later passes check this gradient-presence pattern and unchanged parameter checksums. The score of rollout $i$ differentiates $\sum_t m_{it}\log\pi_\theta(y_{it}\mid\text{prefix})$, retaining policy tokens and masking tool results. Scores use one forward/backward pass per rollout, are stored in single precision, and feed double-precision Gram products and weighted-vector sums. No random projection is used.

\paragraph{Estimators.} MC uses $a_i=(R_i-\bar R)/G$. SW uses $a_i=A_i\rho_i/T_{\rm ref}$ with group-standardized $A_i$ and the recorded importance weight $\rho_i$ (zero when masked). In a full checkpoint probe,
\[
T_{\rm ref}=24\cdot8\cdot\frac{1}{1536}\sum_{i=1}^{1536}T_i,
\]
where $T_i$ counts loss tokens across the 512 clean and 1{,}024 diagnostic rollouts. This constant is shared across groups and designs. It approximates a DAPO loss-gradient scale but is not each training batch's random token count. A deterministic common factor cancels from the variance ratio; a random denominator need not. Neither estimator includes Adam or gradient clipping. Group norms use $a^\top K a$ and pooled means use running weighted-vector sums.

\paragraph{Outcome noise.} For each clean group with $k$ successes, independent masks have probabilities $(1-q)^{\rm kept}q^{\rm flipped}$ and paired masks keep or flip all successes with probabilities $1-q$ and $q$. Advantages are recomputed for each mask; $k=0$ contributes zero. Exactness concerns grader corruption conditional on 64 sampled clean groups, not the population of tasks and trajectories. The between-group term includes the sampling variability of the one clean group available per task.

\paragraph{Transition noise.} Each diagnostic task has eight rollouts under each of eight schedules. Paired groups use a single schedule; 64 independent-comparator groups sample schedule labels with replacement and distinct available rollouts within each label; 64 stratified groups use one rollout per label. These are finite-pool approximations to fresh draws. Index lists are stored. Pooled traces use $n^{-1}\sum_j\|g_j\|^2-\|n^{-1}\sum_jg_j\|^2$, with the analogous within-task estimate and the between-task remainder. Overlapping groups are not independent replications. Reward contrasts are estimated as twice the average within-group sample reward variance.

\paragraph{Evidence and availability.} Compact evidence records task Gram matrices, rewards, token counts, weights, resampling indices and aggregate vector products; within-task Gram matrices alone cannot recover cross-task inner products. Reported inspection checks reproduce summaries with relative and absolute tolerances both set to $10^{-6}$. The eight checkpoints share a coordinate set and scoring commit, have distinct adapter fingerprints, and reuse the base computation. The code repository supplies numerical summaries, but not raw trajectories or Gram evidence for independent reconstruction.

\paragraph{Trainer-loss validation.} A base-model smoke test compares the negative trainer-loss gradient against $T^{-1}\sum_i A_i\rho_iS_i$ on one mixed clean group of size two using its actual token count. The recorded cosine is 0.99994 and relative $\ell_2$ difference 0.017. This is close, not exact, agreement between separately evaluated finite-precision computations. It does not validate loaded checkpoints or quantify variance in full training groups with random batch denominators. Full-size probes do not repeat this batched-gradient check.

\begin{table}[p]
\centering
\caption{Outcome probe: $L$, $R$ and MC traces in units of $10^5$; success, masking and nonzero weights over 512 clean rollouts. P: paired; I: independent.}
\label{tab:probe_extra}
{\small\begin{tabular}{@{}l r rrrr rrr@{}}
\toprule
Traj. & step & $L$ & $R$ & trace P & trace I & clean (\%) & mask (\%) & $\bar\rho_{\ne0}$ \\
\midrule
base & 0 & 0.819 & 2.980 & 0.726 & 0.921 & 77.0 & 5.7 & 0.364 \\
B & 20 & 0.735 & 3.031 & 0.652 & 0.858 & 83.2 & 7.2 & 0.390 \\
B & 40 & 0.653 & 3.338 & 0.579 & 0.821 & 86.5 & 5.9 & 0.352 \\
B & 60 & 0.793 & 3.602 & 0.704 & 0.957 & 86.5 & 4.3 & 0.358 \\
B & 80 & 0.697 & 3.383 & 0.618 & 0.860 & 86.5 & 6.4 & 0.345 \\
B & 100 & 0.676 & 3.433 & 0.600 & 0.848 & 86.3 & 3.9 & 0.330 \\
A & 80 & 0.548 & 2.891 & 0.486 & 0.697 & 86.5 & 3.9 & 0.481 \\
A & 100 & 0.596 & 2.754 & 0.528 & 0.722 & 84.8 & 4.9 & 0.440 \\
\bottomrule
\end{tabular}
\par}
\vspace{1.25em}
\caption{Transition probe (MC): traces in units of $10^5$; empirical mean norms and cosine without uncertainty estimates. P: paired; I: independent resampling; S: stratified.}
{\small\begin{tabular}{@{}l r rrr rrr@{}}
\toprule
Traj. & step & trace P & trace I & trace S & $\|\bar g_P\|$ & $\|\bar g_I\|$ & cosine \\
\midrule
base & 0 & 0.634 & 1.055 & 1.148 & 22.9 & 31.1 & 0.704 \\
B & 20 & 0.649 & 1.254 & 1.397 & 23.5 & 34.5 & 0.659 \\
B & 40 & 0.641 & 1.522 & 1.649 & 24.1 & 39.9 & 0.626 \\
B & 60 & 0.628 & 1.624 & 1.830 & 23.3 & 41.2 & 0.626 \\
B & 80 & 0.642 & 1.673 & 1.888 & 24.0 & 47.2 & 0.635 \\
B & 100 & 0.675 & 1.826 & 2.035 & 25.3 & 48.5 & 0.613 \\
A & 80 & 0.569 & 1.465 & 1.642 & 22.7 & 40.0 & 0.667 \\
A & 100 & 0.560 & 1.452 & 1.579 & 21.6 & 39.9 & 0.593 \\
\bottomrule
\end{tabular}
\par}
\vspace{1.25em}
\caption{Independent trace fractions: conditional grader-mask noise ($f_{\rm out}$) and within-task transition variability ($f_{\rm trans}$, including policy randomness). $V_P$, $V_I$: paired and independent transition reward-contrast estimates.}
{\small\begin{tabular}{@{}l r rrrr@{}}
\toprule
Traj. & step & $f_{\rm out}$ & $f_{\rm trans}$ & $V_P$ & $V_I$ \\
\midrule
base & 0 & 0.291 & 0.879 & 0.223 & 0.338 \\
B & 20 & 0.318 & 0.882 & 0.199 & 0.331 \\
B & 40 & 0.366 & 0.880 & 0.151 & 0.322 \\
B & 60 & 0.339 & 0.878 & 0.148 & 0.316 \\
B & 80 & 0.354 & 0.878 & 0.136 & 0.296 \\
B & 100 & 0.364 & 0.875 & 0.148 & 0.327 \\
A & 80 & 0.373 & 0.876 & 0.149 & 0.304 \\
A & 100 & 0.343 & 0.874 & 0.156 & 0.318 \\
\bottomrule
\end{tabular}
\par}
\end{table}

\end{document}